\documentclass[journal]{IEEEtran}
\usepackage{epsfig}
\usepackage{graphicx}
\usepackage{amsmath}
\usepackage{amssymb}
\usepackage{pifont}%

\usepackage{diagbox}
\usepackage{multirow}
\usepackage[normalem]{ulem}
\usepackage[table]{xcolor}
\usepackage{color}
\usepackage{booktabs}
\usepackage{tabularx}
\usepackage{subfigure}

\newcolumntype{C}[1]{>{\centering\arraybackslash}p{#1}}

\usepackage{threeparttable}
\usepackage{makecell}
\usepackage{enumitem}
\usepackage{paralist}
\usepackage{colortbl}
\let\itemize\compactitem
\let\enditemize\endcompactitem
\let\enumerate\compactenum
\let\endenumerate\endcompactenum

\definecolor{mygray}{RGB}{252,227,227}
\usepackage{bbding}         

\newcommand{\ie}{\textit{i}.\textit{e}.}
\newcommand{\eg}{\textit{e}.\textit{g}.}

\usepackage[pagebackref=true,breaklinks=true,letterpaper=true,colorlinks,bookmarks=false]{hyperref}
\ifCLASSINFOpdf
\else
\fi
\begin{document}

\title{DreamCinema: Cinematic Transfer with Free Camera and 3D Character}

\author{ 
        Weiliang~Chen,
        Fangfu~Liu,
        Diankun~Wu,
        Haowen~Sun,
        Jiwen~Lu,~\IEEEmembership{Fellow,~IEEE}
        and~Yueqi~Duan,~\IEEEmembership{Member,~IEEE,}
\thanks{Weiliang Chen, Fangfu Liu, Diankun Wu, and Yueqi Duan are with the Department of Electronic Engineering, Tsinghua
University, Beijing 100084, China (e-mail: cwl24@mails.tsinghua.edu.cn; liuff23@mails.tsinghua.edu.cn; wdk21@mails.tsinghua.edu.cn;  duanyueqi@tsinghua.edu.cn).

Haowen Sun and Jiwen Lu are with the Department of Automation, Tsinghua University, Beijing 100084, China (e-mail: sunhw24@mails.tsinghua.edu.cn;  lujiwen@tsinghua.edu.cn).

Corresponding author: Yueqi Duan
}
}


\maketitle

\begin{abstract}
 We are living in a flourishing era of digital media, where everyone has the potential to become a personal filmmaker. Current research on video generation suggests a promising avenue for controllable film creation in pixel space using Diffusion models. However, the reliance on overly verbose prompts and insufficient focus on cinematic elements (\eg, camera movement) results in videos that lack cinematic quality. Furthermore, the absence of 3D modeling often leads to failures in video generation, such as inconsistent character models at different frames, ultimately hindering the immersive experience for viewers. In this paper, we propose a new framework for film creation, ~\textbf{DreamCinema}, which is designed for user-friendly, 3D space-based film creation with generative models. Specifically, we decompose 3D film creation into four key elements: 3D character, driven motion, camera movement, and environment. We extract the latter three elements from user-specified film shots and generate the 3D character using a generative model based on a provided image. To seamlessly recombine these elements and ensure smooth film creation, we propose structure-guided character animation, shape-aware camera movement optimization, and environment-aware generative refinement. Extensive experiments demonstrate the effectiveness of our method in generating high-quality films with free camera and 3D characters.

\end{abstract}


\IEEEpeerreviewmaketitle

\section{Introduction}

\label{sec:intro}
With the evolution of digital media, a widespread and flourishing need arises for efficiently creating personal, high-quality, cinematic-level videos~\cite{xing2023surveyvideodiffusion, zhou2024surveyvideogeneration}.
However, film creation has always been a process marked by high technical difficulty~\cite{mateer2014digitalcinematech}, extensive time requirements~\cite{chen2023budgetingtime}, and considerable costs~\cite{mckenzie2012budget}, as filmmakers have to find appropriate characters and design intricate cinematography and character motions to enhance expressive effects and craft compelling narratives. Therefore, creators are eagerly pursuing innovative technologies to enable efficient and wallet-friendly film production.

With the recent compelling success of large-scale AIGC techniques~\cite{huang2025spectralar, chen2025scenecompleter,  xu2025keeptip2025, zhu2024hightip2024, sheng2023towardtip2023, hyun2023frequencytip2023, gan2022arbitrarytiptexture}, video generation~\cite{henschel2024streamingt2v, jiang2023videobooth} suggests a potential avenue for efficient film creation, as demonstrated by Sora~\cite{sora2024}, which can produce visually appealing, attention-grabbing videos. However, the videos generated by these methods often fail to maintain visual consistency (\eg, imcomplete character)~\cite{sunsoralimitation, liu2024soralimitation} and defy physical intuition (\eg, exaggerated character movement)~\cite{cho2024soraphys} due to the lack of decoupling of visual elements such as camera and character motions, as well as insufficient 3D character modeling, ultimately failing to fully immerse viewers in the video content. Additionally, because the prompts based on text or images lack rich cinematic knowledge, video generation models struggle to create videos with cinematic quality. Although a significant amount of research~\cite{hu2024animateanyone, wang2024unianimate} has focused on character image animation, which generates video by animating a reference image with a pose sequence. However, these methods commonly suffer from similar limitations because they are fundamentally built upon 2D diffusion models and ControlNet~\cite{zhang2023adding-controlnetpaper}. Therefore, efficiently constructing and integrating visual elements to produce 3D-consistent films with well-designed camera movements remains a crucial challenge. 

\begin{figure*}[!t]
    \centering
    \includegraphics[width=1\linewidth]{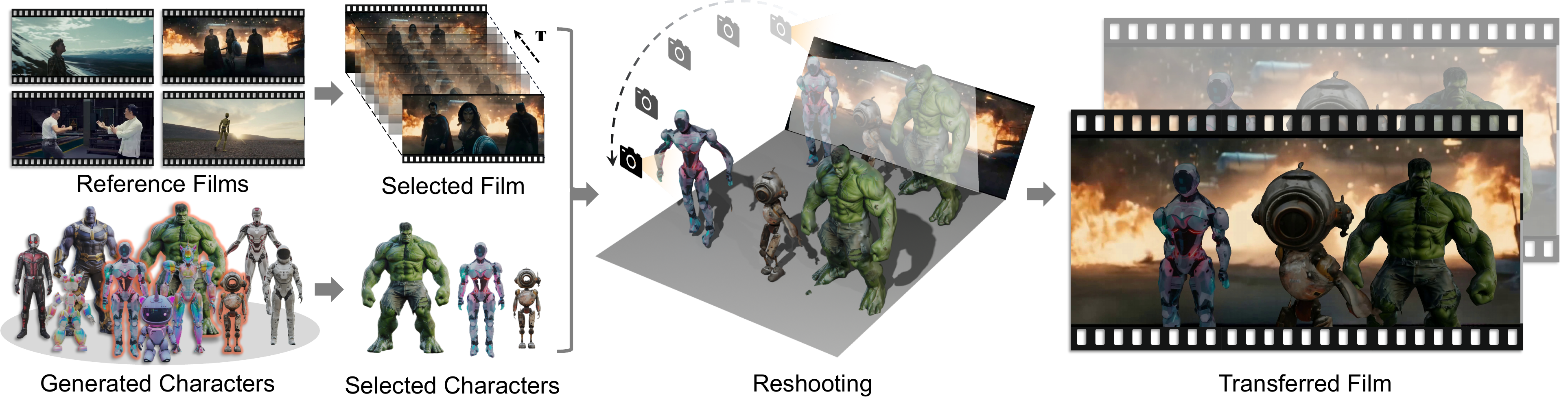}
    \vspace{-6mm}
    \caption{\textbf{DreamCinema} is a user-friendly 3D film creation framework that facilitates personal movie creation with free camera and desired characters. DreamCinema begins by decomposing a 3D film into four key elements: 3D characters, driven motion, camera movement, and environment, all of which are constructed from simple user prompts (i.e., a film shot and a character image). It then recombines these elements smoothly and seamlessly to generate a new film using three well-designed harmonization techniques.}
    \label{fig:teaser}
    \vspace{-6mm}
\end{figure*}

Drawing inspiration from the prevalent watch-and-learn paradigm in film production, recent works~\cite{jiang2023cinematictransfer, wang2023jaws} have focused on cinema behavior transfer, attempting to extract cinematic knowledge from movie scenes for subsequent filmmaking. Leveraging advancements in camera pose estimation~\cite{yen2021inerf, zhu2022nice-slam} and human motion estimation~\cite{pavlakos2019expressivesmplx, ye2023decoupling} technologies, these studies extract visual elements such as camera trajectories and SMPL~\cite{loper2023smpl} tracks from film clips for classic recreations. However, these works primarily focus on the camera movement and driven motion extraction, while overlooking the 3D character, which is a key component of films and typically requires time-consuming and costly manual crafting. Moreover, directly applying the extracted driven motion and camera movement to newly crafted characters often causes disharmony, as these characters have different structures and shapes compared to those in the original shots, making the process less flexible and more impractical. 
Since generative models~\cite{foo2023aigcsurvey, xu2022selftip2022, huo2022casttip2022} have demonstrated remarkable efficiency~\cite{zou2023triplanemeetsgaussian}, high quality~\cite{liu2023sherpa3d, long2023wonder3d} and customizability~\cite{liu2024makeyour3d, ruiz2023dreambooth} across many fields, it naturally prompts us to explore how they can be harnessed to enhance the film production paradigm.

To tackle the above challenges, we propose \textbf{DreamCinema}, a novel cinematic transfer framework in 3D space (shown in Fig.~\ref{fig:teaser}), for user-friendly film creation. Our key insight is to decompose the 3D film into four key components: 3D character, driven motion, camera movement, and environment, and model them in 3D space which naturally preserves 3D consistency and provides flexible manipulation for subsequent creation. We first extract the latter three elements from provided film shots with world-grounded human motion recovery method~\cite{shen2024gvhmr} and video inpainting method~\cite{zhou2023propainter} while adopting an efficient and high-fidelity mesh generation method~\cite{wu2024unique3d} to generate 3D characters. An intuitive approach would be to drive the character with the estimated motion, re-shoot it with the camera, and then integrate the animation with the environment. 

However, the generated videos are both disharmonious and of low quality, which we attribute to the following issues: 1) \textbf{mismatch between the character and original motion}: this occurs because the character specified by the user often has a different structure and shape from the character in the film. Directly animating the character leads to a loss in motion fidelity. 2) \textbf{misalignment between the animated character and camera}: As the estimated camera movement is designed for the original character, it becomes misaligned and fails to capture the details when re-shooting a new animated character. 3) \textbf{disharmony between the re-shot character and environment}: The mismatch in tone, style, and other attributes between the character and environment creates a disjointed effect, often resulting in physical inconsistencies such as unnatural lighting. To address these issues respectively, we propose the following solutions: 1) structure-guided character animation, which aligns the generated character with the driven motion’s canonical skeleton, ensuring motion fidelity; 2) shape-aware camera movement optimization,  which bridges the 3D animation and 2D original shot alignment in SMPL space, enabling accurate shot reproduction during re-shooting, and 3) environment-aware generative refinement, which seamlessly integrate the re-shot animation into the environment with a generative model. Extensive experiments demonstrate the effectiveness of our method in generating high-quality films with free camera and 3D characters.

The main contributions of this work are summarized as follows:
\begin{itemize}
    \item We propose \textbf{DreamCinema}, a novel 3D cinematic transfer framework that decomposes film creation into four orthogonal components—\textit{3D character}, \textit{driven motion}, \textit{camera movement}, and \textit{environment}—and explicitly models them in 3D space. This decoupling enables consistent geometry, flexible editing, and high-quality new film creation from arbitrary inputs.
    
    \item To address misalignment and inconsistency issues, we design a structured, multi-stage system: (i) \textit{structure-guided character animation} that preserves motion fidelity by aligning with the canonical skeleton, (ii) \textit{shape-aware camera movement optimization} for accurate re-shooting in SMPL space, and (iii) \textit{environment-aware generative refinement} that harmonizes lighting and style, ensuring seamless character-environment integration.
    
    \item We conduct comprehensive experiments, including qualitative comparisons, quantitative metrics on 3D consistency and image realism, ablation studies, perturbation tests, and user studies. Results demonstrate that our method significantly outperforms state-of-the-art 2D animation and video editing baselines, particularly in challenging scenarios with dynamic motion and camera movement.
\end{itemize}

\section{Related Work}

\subsection{Character Image Animation.}
Character image animation aims to drive a character image using signals (\eg, videos or skeleton sequences) to generate videos. Recently, with the superior generation capabilities of diffusion models~\cite{croitoru2023visiondiffusion, ho2020ddpm}, many studies~\cite{wang2024unianimateanimation, xu2024magicanimateanimation, ren2020deeptipimageanimation} focus on using video diffusion models~\cite{henschel2024streamingt2v, jiang2023videobooth} for character image animation. DreamPose~\cite{karras2023dreamposeanimationdreampose} adapts the pretrained Stable Diffusion~\cite{blattmann2023stable} into a pose-and-image guided video generation model, fine-tuning it to generate animated fashion video. Disco~\cite{wang2024discoanimationdisco} integrates CLIP~\cite{radford2021learningclippaper} and ControlNet~\cite{zhang2023adding-controlnetpaper} to provide disentangled control for dancing video synthesis. Following this, Animate Anyone~\cite{hu2024animateanyone} and MagicAnimate~\cite{xu2024magicanimateanimation} introduce temporal-attention blocks to enhance temporal consistency, thus generating more coherent videos. However, without 3D modeling, the animations generated lack 3D consistency. Additionally, overlooking the importance of cinematography makes it difficult to produce film-quality videos. Concurrent work like MIMO~\cite{men2024mimo} acknowledges the importance of 3D motion and integrate it as a condition, but it still models other elements in 2D space and thus suffers from issues mentioned above. In contrast, our framework decomposes film shots into four components, and models them in 3D space, thereby achieving 3D-consistent, film-quality video creation.

\subsection{3D Generative Models.}
With the recent success of image~\cite{croitoru2023visiondiffusion, ho2020ddpm, huang2025spectralar} and video~\cite{henschel2024streamingt2v} generation, numerous works~\cite{poole2022dreamfusion, liu2023sherpa3d, liu2024makeyour3d, long2023wonder3d} have focused on utilizing these pretrained 2D diffusion models for 3D generation to address the scarcity of 3D data. Pioneered by DreamFusion~\cite{poole2022dreamfusion}, a series of works~\cite{lin2023magic3dfusion, chen2023fantasia3dfusion, liu2023sherpa3d} have used Score Distillation Sampling (SDS) to achieve 3D generation by distilling different perspectives of 3D models from 2D diffusion models. However, these methods require long per-case optimization and suffer from issues such as multi-face problems due to the lack of 3D priors, which limits their practical application. Building on Zero-1-to-3~\cite{liu2023zero123multiview}, numerous works~\cite{long2023wonder3d, wu2024unique3d, yang2023singletip20233d, lei2022c2fnettip3d} now explore 3D generation following the paradigm of first generating multi-view images via diffusion models fine-tuned on 3D data and then performing sparse view reconstruction, which achieve both 3D consistency and efficiency. Among them, Unique3D~\cite{wu2024unique3d} generates high-quality meshes efficiently with multi-level upscaling strategy and mesh optimization. For this reason, we select it as our 3D character generator. However, how to leverage the generated 3D models for video creation, particularly for producing films with cinematography, is still under exploration. This is the primary focus of our research.

\subsection{World-grounded Human Motion Recovery.} 
World-grounded human motion recovery aims to reconstruct continuous 3D human motion in world coordinates. Previous works~\cite{bogo2016keep, lassner2017unite} focus on recovering motion in camera coordinate system, which requires camera pose for transforming it into world-space. SLAMHR~\cite{ye2023slamhr} combine SLAM~\cite{teed2021droidslam} with 3D human models~\cite{rempe2021humor3dhumanmodelrecovery} to recover world-grounded motion and camera poses via joint optimization. However, the optimization is time-consuming, especially for long videos, which fail to converge. WHAM~\cite{shin2024whamhumanmotionrecovery} autoregressively estimates per-frame poses and translations and still suffers from error accumulation. GVHMR~\cite{shen2024gvhmr} addresses this by estimating human poses per frame in a gravity-view space and transforming them back into world coordinates, achieving efficiency while avoiding error accumulation.
Despite these advancements, how to integrate the estimated motion and camera in subsequent tasks remains an open problem. Recent works like Jaws~\cite{wang2023jaws} and CineTrans~\cite{jiang2023cinematictransfer} recognize the importance of human motion and camera in cinematic transfer. However, they still focus on aligning the camera-rendered motion with the original shot, which reduces their task to world-grounded human motion recovery. In contrast, we rethink the potential problems (\eg, mismatches between animated character and camera movement) during transfer, proposing that optimization should focus on the new character and shot to improve the overall quality of the newly created film.

\section{Method}
\begin{figure*}[!t]
    \centering
    \includegraphics[width=1\linewidth]{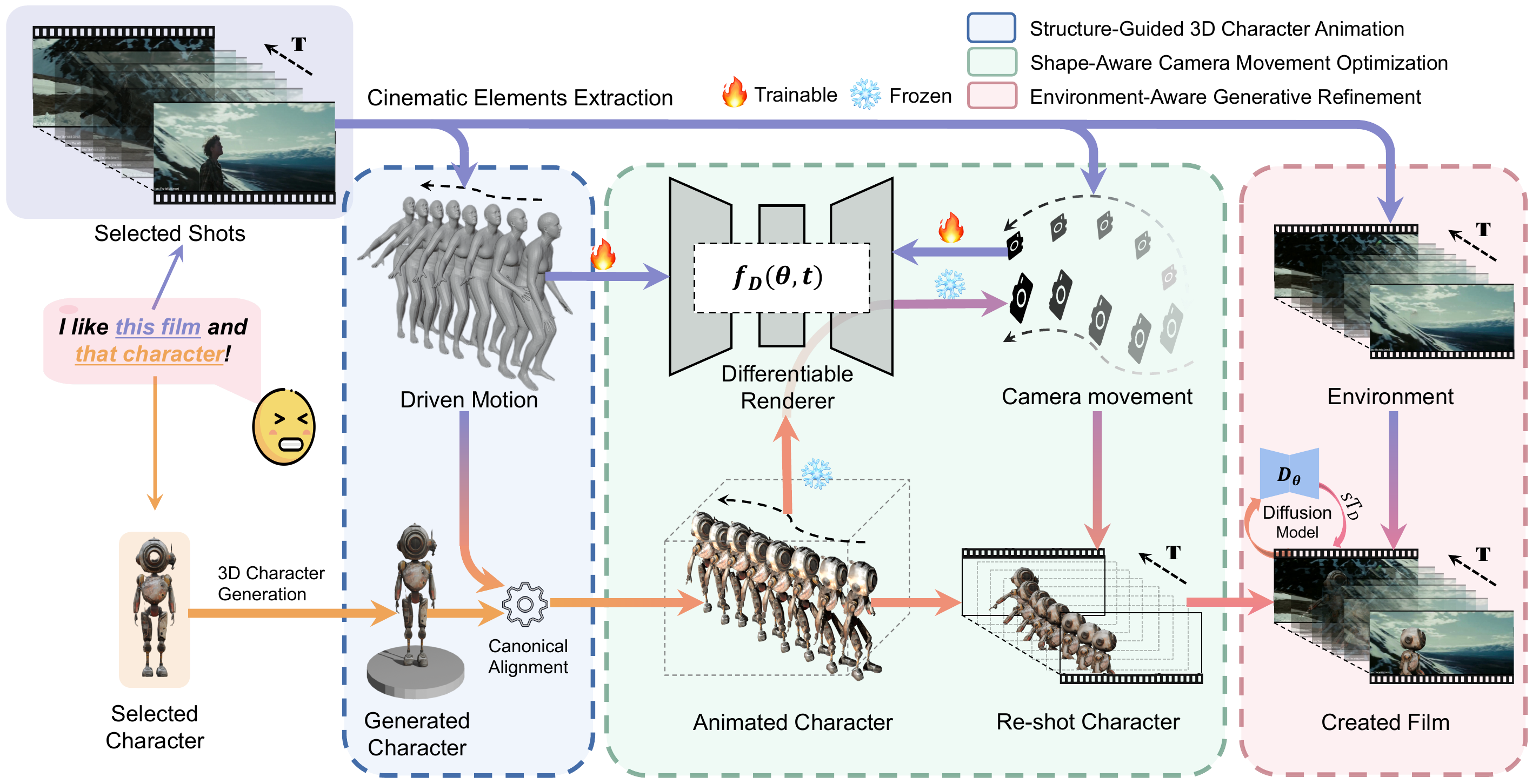}
    \vspace{-6mm}
    \caption{\textbf{The overall framework of DreamCinema.} We first extract the cinematic elements (\ie 3D character, driven motion, camera movement and environment) from the reference shot and image (Sec.~\ref{extraction}). Next, we animate the generated 3D character with our Structure-Guided 3D Character Animation method (Sec.~\ref{animation}). As the animated character misaligns with the original camera movement, we propose Shape-Aware Camera Movement Optimization with a differentiable renderer to achieve seamlessly re-shooting (Sec.~\ref{optimization}). Finally, we design the Environment-Aware Generative Refinement by leveraging a diffusion model to improve overall performance (Sec.~\ref{refinement}). Our framework can create novel films with generated elements tailored to user preference.}
    \label{fig:pipeline}
    \vspace{-6mm}
\end{figure*}
\label{sec:method}
In this section, we introduce our DreamCinema, a user-friendly cinematic transfer framework with a free camera and 3D characters. Our goal is to construct the four key components in film creation (\ie, 3D character, driven motion, camera movement, environment) from user-specific shots and images and seamlessly recombine them to create novel films powered by AIGC. Firstly, we extract cinematic elements (\ie, driven motion, camera movement, environment) from the selected shot and generate 3D character from the provided image (Sec.~\ref{extraction}). To seamlessly incorporate the four components to create new films, we devise structure-guided 3D character animation (Sec.~\ref{animation}), shape-aware camera movement optimization (Sec.~\ref{optimization}) and environment-aware generative refinement (Sec.~\ref{refinement}), each designed to address: 1) the mismatch between the character and original motion, 2) the misalignment between the animated character and camera movement, and 3) the disharmony between the re-shot character and environment during re-shooting. Before introducing our DreamCinema in detail, we first preview some preliminaries (Sec. ~\ref{preliminaries}). An overview of our framework is depicted in Fig.~\ref{fig:pipeline}.

\subsection{Preliminaries} \label{preliminaries}
\textbf{Diffusion Model.} 
Diffusion models~\cite{ho2020ddpm, song2020denoisingDDIM} generate samples from a Gaussian distribution with two processes: (a) a forward diffusion process that adds noise to the data; (b) a reverse diffusion process that removes the noise to recover the original data distribution. Let \( \mathbf{x}_0 \sim p(\mathbf{x}) \) be the sampled data, and \(\mathbf{c}\) refer to the additional condition (\eg text or image). In training, the model adds noise to \( \mathbf{x}_0 \) over \( T \) time steps using a noising schedule \(\alpha_t \in (0, 1)\), with \(\hat{\alpha}_t = \prod_{s=1}^{t} \alpha_s\). This is formulated as:
\begin{equation}
    \mathbf{x}_t = \sqrt{\hat{\alpha}_t} \mathbf{x}_0 + \sqrt{1 - \hat{\alpha}_t} \boldsymbol{\epsilon},
\end{equation}
where \(\boldsymbol{\epsilon} \sim \mathcal{N}(\mathbf{0}, \mathbf{I})\) is the added noise. The model then learns to estimate the noise given a condition \( \mathbf{c} \) by minimizing the following objective:
\begin{equation}
    \mathcal{L}_{\text{simple}} = \mathbb{E}_{\mathbf{x}_0, t, \boldsymbol{\epsilon}, \mathbf{c}} \left[ \|\boldsymbol{\epsilon} - \boldsymbol{\epsilon}_\theta(\mathbf{x}_t, t, \mathbf{c})\|^2 \right].
\end{equation}
In the inference, the model can use the reverse diffusion process, conditioned on \( \mathbf{c} \), to recover the original data distribution from Gaussian noise.

\textbf{SMPL-X.} SMPL-X~\cite{pavlakos2019expressivesmplx} is a unified 3D model of the human body, which extends SMPL~\cite{loper2023smpl} with fully articulated hands and an expressive face. It contains 10475 vertices and 54 keypoints. SMPL-X is defined by a function  $M(\beta, \theta, \psi)$ parameterized by pose parameters $\theta$ (consists of body pose $\theta_b$, jaw pose $\theta_f$, and finger pose $\theta_h$), shape parameters $\beta$, and expression parameters $\psi$. More formally:
\begin{align}
\label{eq:smplx}
\begin{aligned}
    T(\beta, \theta, \psi) &= \bar{T} + B_s(\beta) + B_p(\theta) + B_e(\psi), \\
    M(\beta, \theta, \psi) &= \mathtt{LBS}(T(\beta, \theta, \psi), J(\beta), \theta, \mathcal{W}),
\end{aligned}
\end{align}
where $\bar{T}$ is the mean template shape; $B_s$, $B_p$, and $B_e$ are the blend shape functions for shape, pose, and expression, respectively; $T(\beta, \theta, \psi)$ is the non-rigid deformation from $\bar{T}$; $M(\beta, \theta, \psi)$ is posed mesh transformed from $T(\beta, \theta, \psi)$ using the linear blend skinning algorithm $\mathtt{LBS}(\cdot)$~\cite{lewis2023poselbs} based on the skeleton joints $J(\beta)$, the target pose $\theta$ and the blend weights $\mathcal{W}$ defined on each vertice.

\subsection{Cinematic Elements Extraction} \label{extraction}
Our ultimate goal is to create new 3D films tailored to user preferences. Therefore, the first step is to decompose and extract the four key components from the given prompts (\ie, a film shot and an image). Here, we introduce the methods we use and explain their advantages. Formally, we define ${\{\mathcal{I}^{t}\}}_{t=1}^T$ and $\mathcal{I}{c}$ as the film shots and images provided by the user, where $T$ is the total number of frames. Let ${\{\mathcal{V}^{n} \in \mathcal{R}^{3}\}}_{n=1}^{\mathcal{N}_{m}}$, ${\{\mathcal{I}^{t}_{P}\}}_{t=1}^T$, ${\{\mathcal{S}_{w}^{t} \in \mathcal{R}^{(N_{s}+1) \times 3}\}}_{t=1}^{T}$, and ${\{\mathcal{C}_{w}^{t} \in \mathcal{R}^{(3+3)}\}}_{t=1}^{T}$ represent the 3D character mesh, pure environment, driven motion, and camera movement, respectively, where $N_{s}$ is the number of joints in the SMPL tracks, and $\mathcal{N}_{m}$ is the number of vertices in the generated mesh.

\textbf{3D Character Generation.}
Since 3D characters in film applications require intricate details and well-defined geometry for better animation, we adopt Unique3D~\cite{wu2024unique3d} for our 3D character generation. Unique3D’s multi-view diffusion and normal diffusion generate multi-view RGB and normal images of the character, resulting in a 3D model with enhanced geometric quality and 3D consistency. Additionally, the multi-level upscaling strategy improves fine details, ensuring that our generated characters maintain excellent visual quality during animation.

\textbf{Driven Motion and Camera Movement Estimation.}
Considering that we create new films with arbitrary driven motion and camera movement, it is crucial to decouple these two elements, which helps prevent the driven motion from becoming distorted or exaggerated under different cinematography. To achieve this, we adopt a world-grounded human motion recovery method introduced in GVHMR~\cite{shen2024gvhmr}, which decouples the driven motion and camera movement using a novel gravity-view coordinate system. Furthermore, the per-frame estimation and global alignment paradigm, compared to other approaches, effectively avoids error accumulation and is more efficient for long video predictions, making it particularly well-suited for our framework. This can be formulated as follows:
\begin{align}
\label{eq:estimation}
\begin{aligned}
    {\{\mathcal{S}_{w}^{t}\}}_{t=1}^{T}, {\{\mathcal{C}_{w}^{t}\}}_{t=1}^{T} = 
    f_{\mathcal{H}}({\{\mathcal{I}^{t}\}}_{t=1}^T)
\end{aligned}
\end{align}
where $f_{\mathcal{H}}$ denotes the GVHMR method~\cite{shen2024gvhmr}.

\textbf{Pure Environment Extraction.}
For the environment, we utilize the state-of-the-art propagation-based and mask-guided video inpainting method, Propainter~\cite{zhou2023propainter}. Combined with Segment Anything Model~\cite{kirillov2023segmentanything}, Propainter effectively tracks foreground objects, discards unnecessary and redundant tokens, and extracts the pure environment.

\subsection{Structure-Guided 3D Character Animation} \label{animation}
As the 3D character is generated from a user-desired image, its body structure (\eg, height, proportions, and joint positions) typically differs from that of the characters in the film shot, thus preventing us from directly applying the motion ${\{\mathcal{S}_{w}^{t}\}}_{t=1}^{T}$ to drive the 3D character ${\{\mathcal{V}^{n}\}}_{n=1}^{\mathcal{N}_{m}}$.
This is the issue we identify as the mismatch between the 3D character and the original motion. Therefore, we propose using the character’s structure to guide character animation. Specifically, we first normalize the 3D character mesh and the canonical skeleton using $\mathcal{L}_{m}$ and $\mathcal{L}_{s}$, which represent the height of the character and the SMPL mesh, respectively, ensuring that the character’s scale aligns. We then extract the key joints from the character's front view using OpenPose~\cite{cao2017realtimeopenpose} as the character’s structure, and calculate the joint angle differences $\Delta\mathcal{R} \in \mathcal{R}^{K}$ between these joints and the canonical skeleton for pose alignment. To address skeletal topology differences and bridge 2D and 3D spaces, we select $K$ key joints from OpenPose and SMPL, construct a bone mapping, and perform 2D pose alignment on the front view of the canonical skeleton. We can then bind the mesh to the canonical skeleton, compensate for the motion using $\Delta \mathcal{R}$ and apply linear blend skinning~\cite{lewis2023poselbs} to animate the character. The entire process can be formulated as follows:
\begin{align}
\begin{aligned}
    {\{\mathcal{M}^{t}\}}_{t=1}^{T} &= \mathtt{LBS}(\frac{\mathcal{L}_{s}}{\mathcal{L}_{m}}{\{\mathcal{V}^{n}\}}_{n=1}^{\mathcal{N}_{m}}, 
    \phi_{\theta}(\{{\mathcal{V}^{n}\}}_{n=1}^{\mathcal{N}_{m}}, {\{\mathcal{S}_{w}^{t}\}}_{t=1}^{T}))
\end{aligned}
\end{align}
where $\phi_{\theta}(\cdot, \cdot)$ returns the adjusted canonical skeleton, the compensated motion, and skinning weights for each vertex, and ${\{\mathcal{M}^{t}\}}_{t=1}^{T}$ denotes the animated character meshes.

\subsection{Shape-Aware Camera Movement Optimization} \label{optimization}
Considering that the camera movement in the film shots is personalized to highlight the character's actions, directly applying the estimated camera movement to re-shoot the animated character may result in a suboptimal cinematic effect. This is primarily due to the adjustments made to the canonical skeleton and driven motion when animating the 3D character, which causes the animated character’s motion to misaligned with the original character and appear unnatural. Therefore, we propose our shape-aware camera movement optimization. Inspired by iNeRF~\cite{yen2021inerf} and CineTrans~\cite{jiang2023cinematictransfer}, we utilize an inverse NeRF optimization process to optimize the camera parameters, aiming to better re-shoot the animated character. The key insight is to bridge the 2D motion in film shots and the 3D character motion in the SMPL space, thereby selecting the most suitable perspective to re-shoot the 3D character. Formally, we first train a NeRF model, denoted as $f_{D}(\theta, t)$, using the adjusted SMPL tracks ${\{\mathcal{\hat{S}}_{w}^{t}\}}_{t=1}^{T}$. Next, we query the trained NeRF with the camera movement and optimize it by leveraging the SMPL mask, keypoints, and motion flow from 2D shots. This process can be formulated as follows:
\begin{align}
\begin{aligned}
    {\{\mathcal{\hat{C}}_{w}^{t}\}}_{t=1}^{T} = \{\arg \min_{\mathcal{\hat{C}}_{w}^{t}} \Sigma_{j}\mathcal{L}_{j}(f_{D}(\mathcal{\hat{C}}_{w}^{t}, t), \hat{I}_{j}^{t})\}_{t=1}^{T}
\end{aligned}
\end{align}
where ${\{\mathcal{\hat{C}}_{w}^{t}\}}_{t=1}^{T}$ represents the optimized camera movement and $\mathcal{L}_{j} \in \{\mathcal{L}_{i}, \mathcal{L}_{s}, \mathcal{L}_{m}\}$ are the instance, semantic, and motion losses, respectively, while $\hat{I}_{j}^{t} \in \{\hat{I}_{i}^{t}, \hat{I}_{s}^{t}, \hat{I}_{m}^{t}\}$ are the corresponding supervisory signals from original shots.

\subsection{Environment-Aware Generative Refinement} \label{refinement}
\begin{figure}[t]
  \centering
  \includegraphics[width=1.0\linewidth]{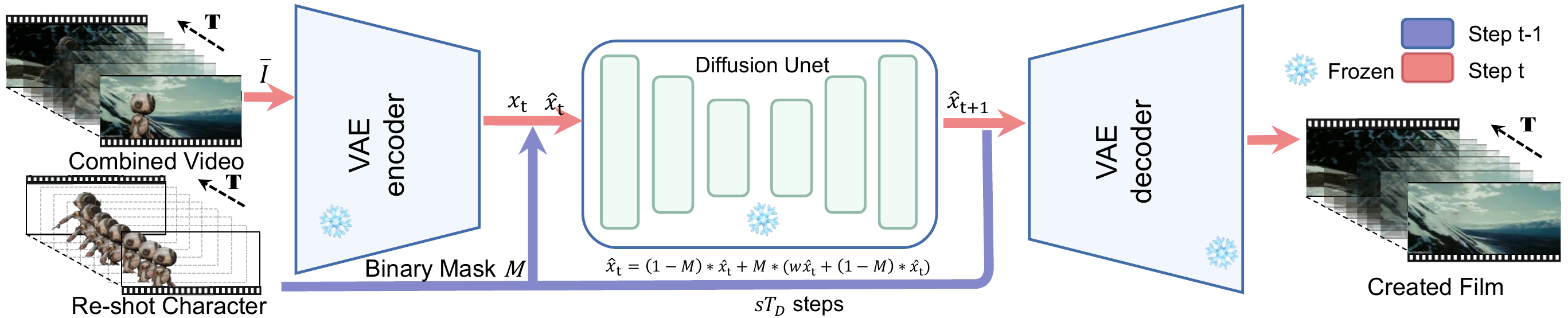}
  \vspace{-6mm}
    \caption{\textbf{Pipeline of environment-aware generative refinement.} We refine the composed video using a diffusion model to address disharmony between the re-shot character and the environment. The process starts from an intermediate noise step and applies latent updates to preserve character appearance while improving overall visual consistency.}
\vspace{-6mm}
   \label{fig:pipeline-generative}
\end{figure}

As the film shot and the character are arbitrarily selected by the user, there is often a significant domain gap between the re-shot character and the environment (\eg, a robot dancing in La La Land). Moreover, as the animated character is re-shot without the environment, this can result in lighting and other environmental factors that do not conform to physical laws, leading to what we refer to as the disharmony between the re-shot character and the environment. Given the powerful ability of diffusion models to generate video from noise, this prompts us to treat our combined video as a strong condition with noise, which we then refine using a generative model, as shown in Fig.~\ref{fig:pipeline-generative}. Formally, we denote our combined video as ${\{\mathcal{\overline{I}}^{t}\}}_{t=1}^T$, and the generative model as $\mathcal{D}_{\phi}$. 
Since the combined video already exhibits high quality, sparked by SDEdit~\cite{meng2021sdeditgraduallydenoise} and PhysGen~\cite{liu2025physgen}, we define a noise strength $s \in [0, 1]$, where the denoising process begins at step $s \times \mathcal{T}_\mathcal{D}$, with $\mathcal{T}_\mathcal{D}$ representing the total steps of the complete denoising process. Additionally, since the animated 3D character is high-quality, we aim to preserve as many of its features as possible. To achieve this, we introduce a latent update weight $w$ for the character. Consequently, the latent is updated at each denoising step as follows:
\begin{align}
\begin{aligned}
    \hat{x}_{t} = (\mathcal{I}-\mathcal{M}) \cdot \hat{x}_{t} + \mathcal{M} \cdot (w \cdot \hat{x}_{t} + (\mathcal{I}-w) \cdot {x}_{t})
\end{aligned}
\end{align}
where ${x}_{t}$ is the noisy reference video in latent space at step $t$, $\hat{x}_{t}$ is the denoised output at step $t+1$, and $\mathcal{M}$ is a binary mask for the animated character.

\begin{figure*}[!t]
    \centering
    \includegraphics[width=1\linewidth]{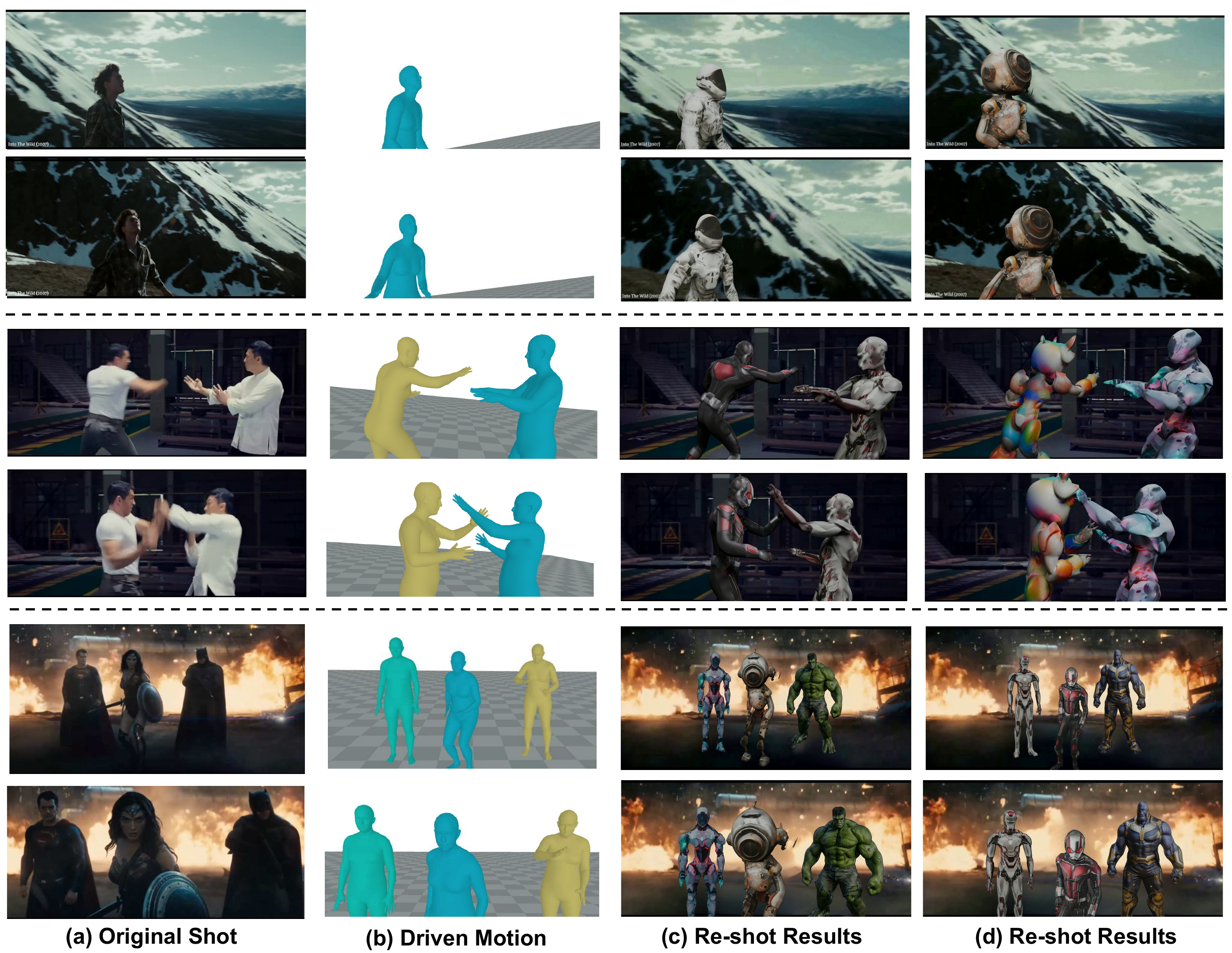}
    \vspace{-6mm}
    \caption{\textbf{Examples of cinematic transfer results.} (a) The original shots. We present three common shot types: Arc, Track, and Push In. (b) Driven motion visualization of our method. We render the extracted driven motion with the optimized camera movement to visualize our extracted cinematic elements. (c)-(d) The re-shot results visualizatioon. We generate diverse characters with high quality and alignment to user preferences and selectively transfer these cinematic elements (\eg cinematography and character motions) to create new films. }
    \label{fig:main_results}
    \vspace{-6mm}
\end{figure*}
\section{Experiments}
\label{sec:exp}

\subsection{Implementation Details}
\label{subsec: implementation details}
Our implementation is based on PyTorch. We utilize GVHMR~\cite{shen2024gvhmr} for driven motion and camera movement estimation. For 3D character generation, we adopt Unique3D~\cite{wu2024unique3d} to produce high-quality, intricate meshes. Combining SAM~\cite{kirillov2023segmentanything} and Propainter~\cite{zhou2023propainter}, we isolate the pure environment from shots. We employ Blender’s~\cite{cushman2011openriggingblender} automatic weight assignment for rigging the character with the adapted motion, and apply linear blend skinning~\cite{kavan2007lbskinning} for animation. For camera movement optimization, we choose D-NeRF~\cite{pons2021dnerf} as the differential renderer. Finally, for generative refinement, we use SEINE~\cite{chen2023seine} as the denoiser and set the $s=0.2$ and $w=0.1$, respectively.

\subsection{Cinematic Transfer Results}
Fig.~\ref{fig:main_results} showcases the results of our cinematic video creation with free film shots and characters. By decomposing film creation into four key components—3D character, driven motion, camera movement, and environment—and modeling them in 3D space, followed by separate optimization at each stage, our framework \textbf{DreamCinema} can generate videos with the following advantages: 
\begin{itemize}
    \item \textbf{3D consistency}: By representing characters in 3D space throughout the pipeline, we ensure that the generated character retains spatial coherence across different views and motions. This avoids typical artifacts such as temporal flickering or inconsistent geometry that are often observed in 2D generation methods.
    \item \textbf{High-fidelity motion}: Our structure-aware character animation module enables precise retargeting of motion from the original character to a new, user-specified 3D character. By aligning the motion to a canonical skeleton and adjusting for structural discrepancies, the animated character exhibits smooth, realistic, and physically plausible motion, faithfully preserving the dynamics of the original film clip.
    \item  \textbf{Diverse camera movements}: As demonstrated with original shots featuring varying camera movements, the results show that our framework can adapt to different cinematic styles. This is made possible by our shape-aware camera movement optimization. 
    \item \textbf{Overall harmony} (\eg, tone, lighting, etc.): Thanks to our environment-aware generative refinement, the generated video maintains a consistent and harmonious feel. 
\end{itemize}

In conclusion, our novel framework \textbf{DreamCinema} offers a unified solution for high-quality video generation by decomposing the filmmaking process into structured sub-problems, each handled with specialized 3D modeling and optimization strategies. This not only significantly improves the visual fidelity and coherence of the generated results but also offers a new perspective on AI-assisted filmmaking—enabling users to flexibly recombine characters, motions, camera styles, and environments to create entirely new cinematic experiences.

\begin{figure}[!t]
    \centering
    \includegraphics[width=1\linewidth]{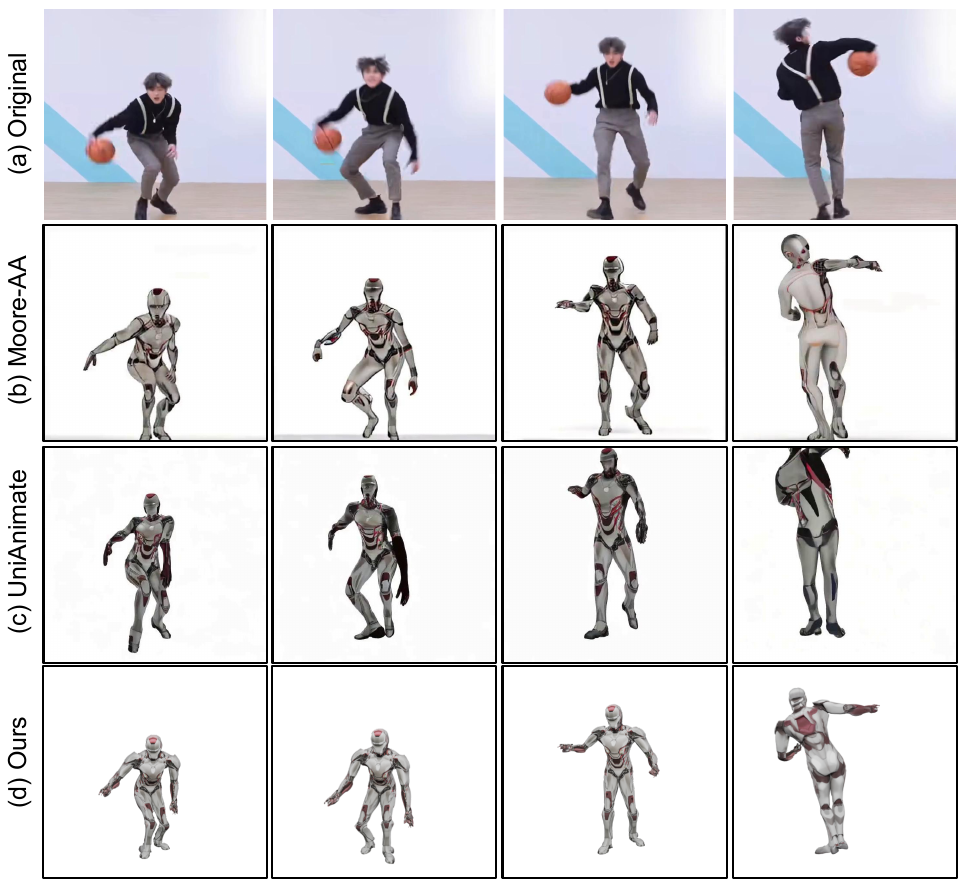}
    \vspace{-6mm}
    \caption{\textbf{Qualitative Comparison with SOTA Character Animation Methods.} We compare our method with Animate Anyone and UniAnimate under a unified background. By removing our environment module, we highlight the superiority of our 3D-based animation in preserving structural integrity and motion coherence during complex movements.}
    \vspace{-6mm}
    \label{fig:comparison}
\end{figure}

\subsection{Comparison with SOTA Character Animation Methods}
\textbf{Baselines.}
As character image animation is most closely related to our work,  we compare our approach with two recent state-of-the-art methods: \textbf{Animate Anyone}\cite{hu2024animateanyone, moore2024animate} and \textbf{UniAnimate}\cite{wang2024unianimate}. Animate Anyone~\cite{hu2024animateanyone} animates arbitrary characters from a single image using 2D pose sequences, achieving high-quality appearance preservation. UniAnimate~\cite{wang2024unianimate} further generalizes human animation through a unified framework that disentangles pose, motion, and appearance. However, both methods operate purely in 2D space, making it difficult to maintain geometric consistency under complex motions and camera movements. In contrast, our method explicitly models characters and motion in 3D space, enabling accurate motion retargeting, realistic camera alignment, and physically plausible character-environment integration.

\textbf{Qualitative Comparison.}
As shown in Fig.~\ref{fig:comparison}, we present a qualitative comparison between our method and two representative 2D-based character animation approaches—Animate Anyone and UniAnimate. To enable a direct and fair comparison focused purely on character animation quality, we remove our environment module and evaluate all methods under a consistent visual background. It is important to note, however, that our full pipeline is designed for the broader task of new film creation, where environment plays a crucial role. Our decomposition-and-refinement framework explicitly models and integrates four key components—character, motion, camera, and environment—allowing greater flexibility and realism in film composition. In contrast, existing character animation methods largely neglect the importance of environment in cinematic generation, limiting their ability to produce coherent and immersive video content.
From the results, we observe that in the videos generated by the 2D-based methods, the character often loses structural integrity and visual identity during intense or articulated movements (e.g., standing up or turning), primarily due to the absence of explicit 3D modeling. Furthermore, these methods frequently produce unnatural distortions and deformations—especially in limb regions like the arms—resulting in perceptually implausible animations. In contrast, our method leverages explicit 3D modeling to consistently preserve the structural integrity and identity of the animated character, regardless of pose complexity or motion intensity, ensuring coherent appearance across complex and temporally extended motions.

\textbf{Quantitative Comparison in 3D Consistency.}  
To comprehensively evaluate the 3D consistency and motion quality of our method, we conduct a quantitative comparison against two state-of-the-art 2D-based character animation approaches: \textbf{Animate Anyone}~\cite{hu2024animateanyone} and \textbf{UniAnimate}~\cite{wang2024unianimate}. The evaluation is performed across various types of cinematic camera shots, including \textit{STATIC}, \textit{PAN}, \textit{DOLLY}, and \textit{ARC}, which involve different levels of complexity in both camera and character motion. We adopt the following three metrics to measure performance from multiple perspectives:  
1) \textbf{Mean Per Joint Position Error (MPJPE)}, which quantifies the average distance between predicted and ground-truth 3D joint positions, directly reflecting the structural consistency and accuracy of character motion in 3D space;  
2) \textbf{Pixel Accuracy (PA)}, which evaluates the overlap between the projected character mask and ground-truth silhouette in the image plane, serving as an indicator of motion fidelity and body pose alignment;  
3) \textbf{Intersection over Union (IoU)}, which measures the spatial alignment between the rendered character and the camera trajectory over time, reflecting the consistency of character-camera interaction.
Since Animate Anyone and UniAnimate are inherently 2D-based and do not provide explicit 3D representations, we extract 3D SMPL motion and camera trajectories from their generated videos using the recent GVHMR~\cite{shen2024gvhmr} model, ensuring a fair and unified evaluation protocol across all methods.

As shown in Table~\ref{tab:comparison}, our method achieves superior performance across all metrics and shot types. The improvement is especially pronounced in dynamic shot types such as \textit{PAN} and \textit{ARC}, where both the character and camera undergo significant motion. These scenarios are particularly challenging for 2D-based methods, which lack a coherent spatial representation and often suffer from geometric inconsistencies or temporal jitter. In contrast, our method explicitly models the character, driven motion, and camera trajectory in 3D space, enabling precise motion retargeting, stable animation, and accurate re-shooting with realistic camera behavior.
Overall, these results demonstrate the strength of our decomposition-and-refinement framework and validate the importance of incorporating explicit 3D modeling in cinematic character animation.

\begin{table*}[t!]
\centering
\caption{\textbf{Quantitative Comparison on 3D Consistency.} We compare our method with Animate Anyone~\cite{hu2024animateanyone} and UniAnimate~\cite{wang2024unianimate} across different shot movement types using MPJPE, PA, and IoU. Our method consistently outperforms the baselines, especially in dynamic shots such as PAN and ARC, validating the effectiveness of our 3D decomposition framework.}

\begin{tabular}{lccccccccc}
\toprule

 \multirow{2}{*}{Methods}  & \multicolumn{3}{c}{PUSH-IN} & \multicolumn{3}{c}{PULL-OUT} & \multicolumn{3}{c}{PAN} \\ 
\cmidrule(lr){2-4}
\cmidrule(lr){5-7}
\cmidrule(lr){8-10}

 & ~~PA$\uparrow$~~      & ~~IoU$\uparrow$~~      & ~MPJPE$\downarrow$~  & ~~PA$\uparrow$~~       & ~~IoU$\uparrow$~~      & ~MPJPE$\downarrow$~    & ~~PA$\uparrow$~~     & ~~IoU$\uparrow$~~    & ~MPJPE$\downarrow$~   \\ 
\midrule

Moore-AA~\cite{moore2024animate} & 78.3   & 84.2   & 83.9      & 77.8   & 83.4   & 83.7      & 73.3 & 72.1 & 409.2     \\

UniAnimate~\cite{wang2024unianimate}     & 80.1   & 76.3   & 82.7      & 81.3   & 80.2   & 81.3       & 71.4 & 70.1 & 483.3     \\

Ours       & \textbf{90.4}   & \textbf{93.3}   & \textbf{58.2}      & \textbf{94.9}   & \textbf{94.5}   & \textbf{57.5}       & \textbf{94.8} & \textbf{93.7} & \textbf{62.2}      \\
\bottomrule 

\multirow{2}{*}{Methods} & \multicolumn{3}{c}{TRACK}   & \multicolumn{3}{c}{FOLLOW}   & \multicolumn{3}{c}{ARC} \\ 
\cmidrule(lr){2-4}
\cmidrule(lr){5-7}
\cmidrule(lr){8-10}

   & ~~PA$\uparrow$~~      & ~~IoU$\uparrow$~~      & ~MPJPE$\downarrow$~  & ~~PA$\uparrow$~~       & ~~IoU$\uparrow$~~      & ~MPJPE$\downarrow$~    & ~~PA$\uparrow$~~     & ~~IoU$\uparrow$~~    & ~MPJPE$\downarrow$~   \\ 
\midrule

Moore-AA~\cite{moore2024animate} & 77.2   & 78.1   & 109.2      & 73.3   & 75.5   & 237.1       & 70.2 & 68.1 & 527.3     \\

UniAnimate~\cite{wang2024unianimate}      & 79.3   & 76.9   & 128.1     & 74.2   & 73.9   & 267.5       & 66.9 & 68.7 & 469.1     \\

Ours       & \textbf{95.0}   & \textbf{94.6}   & \textbf{55.1}      & \textbf{92.4}   & \textbf{90.7}   &  \textbf{61.7}      & \textbf{95.2} & \textbf{94.9} & \textbf{66.3}  \\ 

\bottomrule

\end{tabular}
\vspace{-6mm}
\label{tab:comparison}
\end{table*}

\textbf{Quantitative Comparison in Visual Realism.}
We follow Vbench~\cite{huang2024vbench} to evaluate image and aesthetic quality, as ground truth is unavailable and metrics like FID and LPIPS cannot be used. For image quality (IQ), we use the MUSIQ~\cite{Ke2021MUSIQ} predictor to measure low-level distortions (e.g., noise, blur) frame-by-frame, normalizing scores to [0,1] and averaging across frames. For aesthetic quality (AQ), we adopt the LAION aesthetic predictor~\cite{LAIONaes}, which scores each frame on a 0-10 scale; these are normalized and averaged similarly. We report both metrics as percentage scores for easier interpretation.
As shown in Table~\ref{tab:image_metrics}, our method achieves significantly higher IQ and AQ percentages compared to the baseline, indicating superior visual fidelity and artistic quality. The increased IQ score reflects a substantial reduction in low-level artifacts such as blur, noise, and over-exposure, which are critical for producing clear and stable video frames. Meanwhile, the higher AQ score demonstrates improved overall aesthetics, including better color harmony, composition, and photographic quality. These quantitative improvements align well with our qualitative observations, confirming that our method effectively enhances the realism and visual appeal of generated videos by seamlessly integrating the animated character with its environment.

\begin{table}[h]
    \centering
    \caption{\textbf{Quantitative Comparison in Visual Realism with SOTA Character Animation Methods.} We report image quality (IQ) and aesthetic quality (AQ) scores following VBench~\cite{huang2024vbench}. Our method achieves higher scores than baselines, demonstrating superior visual fidelity and artistic appeal.}

    \begin{tabular}{cccc}
\toprule

                                        & Moore-AA~\cite{hu2024animateanyone} & UniAnimate~\cite{wang2024unianimate}   & ~~~~Ours~~~~   \\ \hline
\multicolumn{1}{c}{AQ$\uparrow$}                                             & 42.5 & 39.6 &    \textbf{53.8}\\
\multicolumn{1}{c}{IQ$\uparrow$}                                             & 43.7 & 44.3 &      \textbf{60.2}\\
\bottomrule
\end{tabular}
\label{tab:image_metrics}
\end{table}

\begin{figure}[h]
  \centering
  \includegraphics[width=1.0\linewidth]{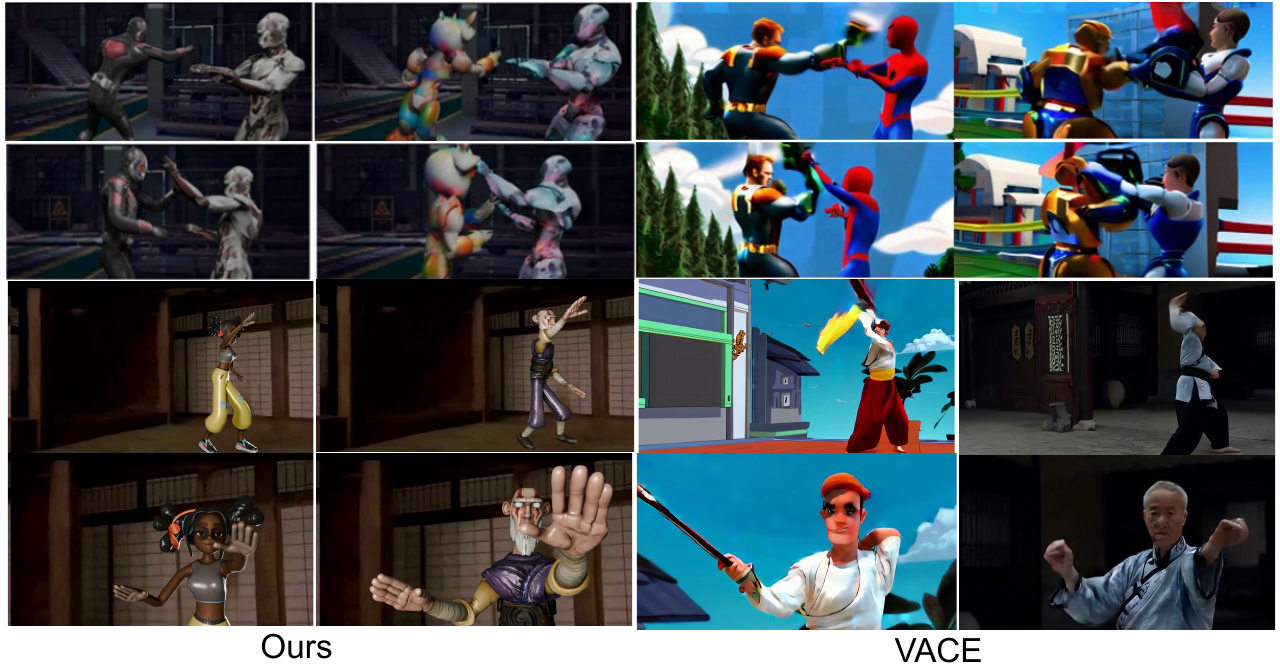}
  \vspace{-6mm}
   \caption{\textbf{Comparison with SOTA Video Editing Method.} Compared to the end-to-end video editing method VACE~\cite{jiang2025vaceediting}, our 3D-based multi-stage framework achieves better motion fidelity, character consistency, and spatial coherence by explicitly modeling four key components in 3D space.}
   \label{fig:comparisonsota}
   \vspace{-6mm}
\end{figure}

\begin{table}[h]
    \centering
    \caption{\textbf{Quantitative Comparison in Visual Realism with SOTA Video Editing Methods.} Our method outperforms VACE~\cite{jiang2025vaceediting} in  visual realism, confirming that explicit 3D modeling of key elements leads to better spatial consistency, and overall visual quality in new film creation.}

    \begin{tabular}{ccc}
\toprule
          & ~~~VACE~\cite{jiang2025vaceediting}~~~  & ~~~~~~Ours~~~~~~   \\ 
\midrule
\multicolumn{1}{c}{Aesthetic Quality$\uparrow$}           & 43.7 &    \textbf{53.8}\\
\multicolumn{1}{c}{Image Quality$\uparrow$}             & 46.2 &      \textbf{60.2} \\
\bottomrule
\end{tabular}

\label{tab:image_metrics_videoedting}
\end{table}

\subsection{Comparison with SOTA in Video Editing}
\textbf{Baselines.}
Recently, with the rapid development of video generation models, several video editing approaches (\eg, ConceptMaster~\cite{huang2025conceptmasterediting}, VACE~\cite{jiang2025vaceediting}, FullDiT~\cite{ju2025fullditediting}) have aimed to build end-to-end systems for unified video editing. Given a content input (such as an image or a text prompt) and a reference video, these methods attempt to perform new film creation in a single pass. However, when dealing with challenging scenarios involving complex character motion and dynamic camera movements, these models often fail to produce film shots with coherent 3D consistency. In this work, we compare with VACE~\cite{jiang2025vaceediting}—the only open-sourced method among them—as a representative baseline, to demonstrate the superiority of our framework, which decomposes video elements into \textit{character}, \textit{driven motion}, \textit{camera movement}, and \textit{environment}, and explicitly models them in 3D space.

\textbf{Qualitative and Quantitative Comparison.}
As shown in Fig.~\ref{fig:comparisonsota}, our 3D-based multi-stage framework produces visibly better results in terms of motion fidelity, character consistency, and spatial coherence compared to end-to-end video editing methods such as VACE~\cite{jiang2025vaceediting}. End-to-end models generate the entire video in a single forward pass, which requires simultaneously satisfying multiple constraints—including character motion, appearance preservation, camera dynamics, and background consistency. This often results in entangled representations and degraded quality, especially in challenging cases involving fast or articulated motion.
For example, in fight scenes, end-to-end methods frequently generate irregular arm deformations during rapid movement due to the lack of structural priors. In kung fu scenes, occlusions in 2D space often lead to inconsistent appearance across frames—for instance, the same body part may appear with different textures before and after occlusion, breaking temporal coherence.
In contrast, our approach decomposes video generation into four orthogonal components—character, driven motion, camera movement, and environment—and models each explicitly in 3D space. This design enables our method to maintain structural integrity across frames, avoid motion drift and identity distortion, and better reproduce intended cinematic compositions under complex motions and viewpoints.

Table~\ref{tab:image_metrics_videoedting} further supports these qualitative observations by showing that our method achieves higher scores in both image quality and aesthetic quality compared to VACE. These results indicate that our framework not only improves structural and temporal consistency but also enhances the overall visual realism of the generated videos. Both quantitative and qualitative results further demonstrate that, compared to end-to-end video editing methods, our explicit modeling of key elements in 3D space enables significantly better spatial consistency, motion fidelity, and overall visual realism in complex new film creation scenarios.

\begin{figure}[!t]
    \centering
    \includegraphics[width=1\linewidth]{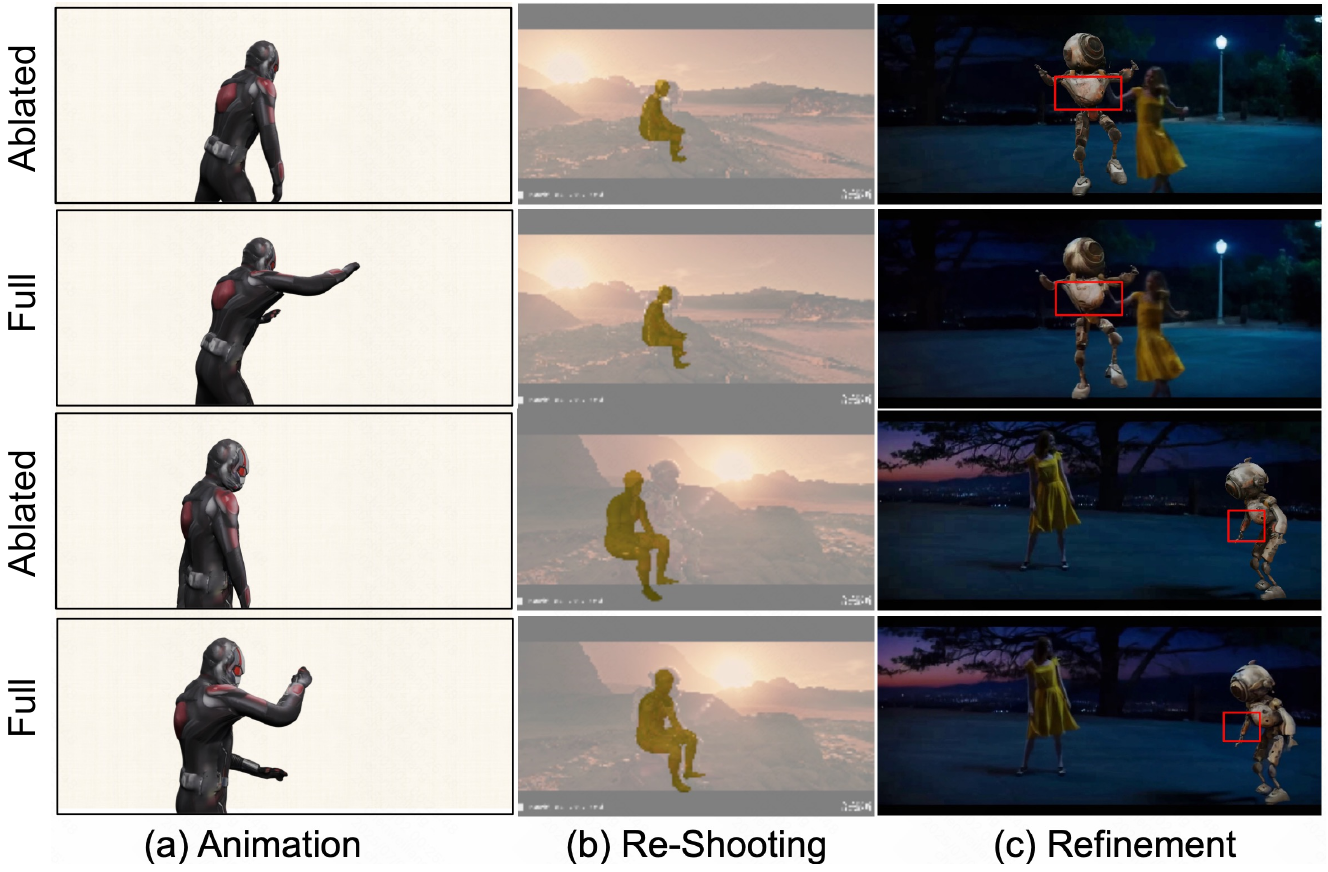}
    \caption{\textbf{Qualitative Ablation Study.} We visualize the impact of each component in our framework. (a) Adaptive animation improves skeletal alignment; (b) camera optimization ensures better motion-camera consistency; (c) generative refinement enhances character-environment integration.}

    \label{fig:ablation}
\end{figure}
\subsection{Ablation Study and Discussion}
In this section, we conduct an ablation study to systematically evaluate the effectiveness of each component in our framework, including structure-guided character animation, shape-aware camera movement optimization, and environment-aware generative refinement. We further analyze the stability of our framework under motion perturbation and discuss the role of generative refinement in enhancing visual realism.

\begin{table}
    \centering
    \caption{\textbf{Quantitative Ablation Study on 3D Consistency.} Adaptive Animation mainly improves PA and IoU, and Camera Optimization reduces MPJPE.}
    \begin{tabular}{cccc}
\toprule
\multicolumn{1}{c}{Methods}                 & ~~~PA$\uparrow$~~~      & ~~~IoU$\uparrow$~~~      & ~~MPJPE$\downarrow$~~ \\ 
\midrule
\multicolumn{1}{c}{w/o Adaptive Animation}                                         &      68.1    &   65.3 &   235.3\\
\multicolumn{1}{c}{w/o Camera Optimization}                                         &    88.2   &   89.3 &   147.1\\
\multicolumn{1}{c}{w/o Generative Refinement}                                         &    93.6    &   \textbf{94.1} &   \textbf{56.1}\\ 
\multicolumn{1}{c}{DreamCinema}                                         &    \textbf{93.8}    &   93.7 &   56.2\\
\bottomrule
\end{tabular}
\label{tab:ablation_study}
\end{table}

\begin{table}[h]
    \centering
    \caption{\textbf{Ablation Study on Image Realism.} Generative Refinement significantly enhances visual quality by improving realism and reducing artifacts.}

    \begin{tabular}{ccccc}
\toprule
             & ~~w/o A-A~~ & ~~w/o C-O~~ & ~~w/o G-R~~ & ~~Full~~      \\ 
\midrule
\multicolumn{1}{c}{AQ$\uparrow$}                      &   46.2 &   48.3 & 49.1 &  \textbf{53.8}\\
\multicolumn{1}{c}{IQ$\uparrow$}                      &   57.9 &   58.2 & 55.8 &  \textbf{60.2}\\
\bottomrule
\end{tabular}
\label{tab:image_metrics_ablation}
\end{table}

\textbf{Overall Ablation Analysis.} As mentioned earlier, directly applying the estimated camera movement and driven motion to the generated character can fail due to inconsistencies and disharmony. We conduct ablation experiments to evaluate our proposed method, and the results are shown in Fig.~\ref{fig:ablation}. Fig.~\ref{fig:ablation} (a) highlights the importance of adaptively matching the structure of the 3D character with the canonical skeleton, which plays a crucial role in assigning skinning weights. Fig.~\ref{fig:ablation} (b) shows that our shape-aware camera movement optimization further aligns the adjusted driven motion with the original shot's character motion. Fig.~\ref{fig:ablation} (c) shows that with our environment-aware generative refinement, the re-shot character integrates more seamlessly into the environment (e.g., in the red-marked area, where the light source is behind the character, and after refinement, the character's shadow appears more natural). 

Quantitative ablation experiments in Table~\ref{tab:ablation_study} further demonstrate the effectiveness of our method in maintaining 3D consistency. As shown in the table, the Adaptive Animation module has the greatest impact on Pixel Accuracy (PA) and Intersection over Union (IoU), indicating its critical role in preserving accurate spatial alignment and character shape. The Camera Movement Optimization module most significantly improves the Mean Per Joint Position Error (MPJPE), reflecting its importance in enhancing precise 3D motion reconstruction. In contrast, the Generative Refinement module contributes less to maintaining 3D consistency metrics. However, as indicated by the ablation results on image realism in Table~\ref{tab:image_metrics_ablation}, Generative Refinement plays a major role in enhancing the visual quality of the generated videos, substantially improving realism and reducing artifacts.

\begin{figure*}[h]
  \centering
  \includegraphics[width=1.0\linewidth]{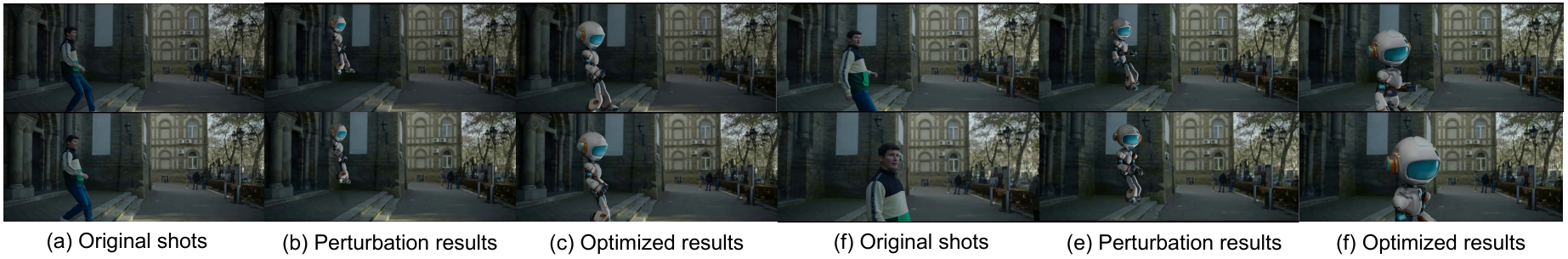}
  \vspace{-6mm}
   \caption{\textbf{Perturbation Experiments on Animation Stability.} By applying Shape-Aware Camera Movement Optimization, our framework adaptively corrects camera parameters to reduce inconsistencies caused by noisy motion extraction, demonstrating robustness in challenging dynamic scenes.}
   \label{fig:perturbation}
   \vspace{-6mm}
\end{figure*}

\textbf{Stability of Our Framework.}
To validate the robustness of our framework, we introduce perturbations to the extracted motion and evaluate its impact on the final output. By applying our Shape-Aware Camera Movement Optimization, we are able to adjust the camera parameters adaptively to better align the character’s joint positions and overall shape in the newly created film with those of the reference video. As shown in Fig.~\ref{fig:perturbation}, this adjustment significantly mitigates inconsistencies caused by noisy or inaccurate motion extraction, particularly in challenging scenarios involving rapid or complex movements. This perturbation experiment demonstrates the stability and resilience of our framework, highlighting its capability to effectively handle errors in motion extraction and still produce coherent and visually consistent new film sequences.

\begin{figure}[h]
  \centering
  \includegraphics[width=1.0\linewidth]{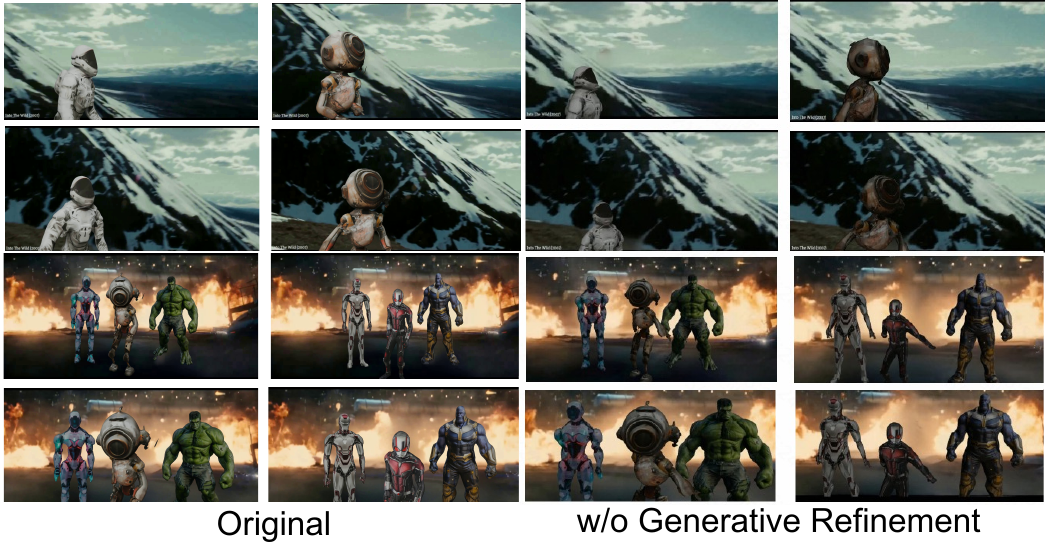}
  \vspace{-6mm}
   \caption{\textbf{Additional Ablation Results on Generative Refinement.} Generative refinement improves character appearance, color consistency, and lighting adaptation, enhancing visual coherence and realism by better integrating the character with the environment.}
   \label{fig:generativeablation}
   \vspace{-6mm}
\end{figure}

\textbf{Discussion of Generative Refinement.}
Fig.~\ref{fig:generativeablation} presents additional ablation results focusing on the impact of generative refinement. Experimental evidence shows that the generative refinement module effectively enhances the character’s appearance, color consistency, and lighting adaptation, enabling the character to better blend with the environment of the reference video. As a result, this refinement leads to the production of videos with improved visual coherence and overall realism, reducing artifacts and discrepancies that arise from direct compositing. These findings demonstrate the critical role of generative refinement in achieving harmonious integration between the re-shot character and complex environmental contexts.

\begin{figure*}[!t]
\centering
\includegraphics[width=1\linewidth]{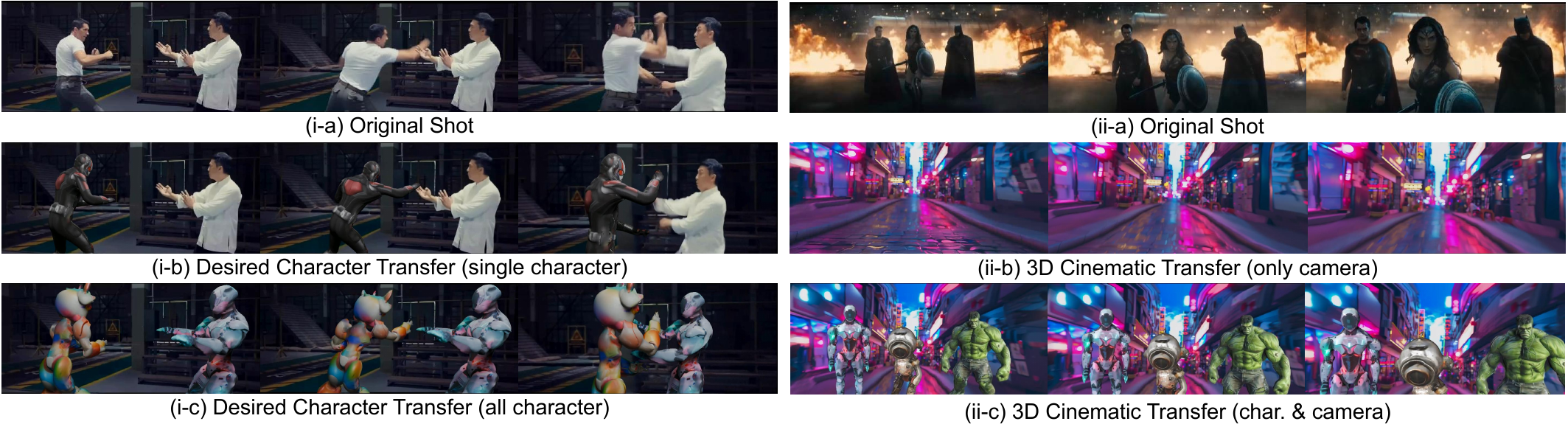}
\vspace{-4mm}
\caption{\textbf{More Flexible Application.} We show more flexible applications of DreamCinema: (i) the classic shots reconstruction with arbitrary characters (ii) new film creation with extracted and generated elements.}
\label{flexible application}
\vspace{-4mm}
\end{figure*}

\subsection{More Applications}
Fig.~\ref{flexible application} shows various applications of our \textbf{DreamCinema}. As shown in Fig.~\ref{flexible application} (i), our framework transfers single or multiple characters, which can be used for classic shot recreation and tribute. The restored shots are highly consistent with the reference shots in character motions, cinematography, and visula aesthetics, enabled by our proposed optimization and refinement. Fig.~\ref{flexible application} (ii) shows our flexible application in creating new films via a 3D engine. Thanks to our paradigm of decomposing the film into four key components, each element can be independently recomposed with new elements. In Fig.~\ref{flexible application} (ii-b) and (ii-c), we show the new film created with extracted camera movement and generated character and new environment. As users can manipulate all the elements arbitrarily via our framework, it demonstrates that our user-friendly \textbf{DreamCinema} has the potential to make everyone to be their own filmmaker.

\begin{table}
    \centering
    \caption{\textbf{User Study.} Comparison on new films creation in seven-point Likert scale (lowest-highest:1-7).
}

    \begin{tabular}{cccc}
\toprule
                  & Moore-AA~\cite{moore2024animate} & UniAnimate~\cite{wang2024unianimate} & ~~~~Ours~~~~ \\ 
\midrule
\multicolumn{1}{c}{CC}          &    3.7$\pm$1.3    &   3.9$\pm$0.8 &   \textbf{6.3$\pm$0.4}\\
\multicolumn{1}{c}{MF}                &    5.4$\pm$1.0    &   5.2$\pm$0.7 &   \textbf{6.5$\pm$0.2}\\
\multicolumn{1}{c}{CMA}              &    3.3$\pm$0.5    &   3.4$\pm$0.4 &   \textbf{6.3$\pm$0.5}\\
\multicolumn{1}{c}{OH}             &    3.8$\pm$1.3    &   4.2$\pm$0.9 &   \textbf{6.2$\pm$0.6}\\
\bottomrule
\end{tabular}

\label{tab:user_study}
\end{table}

\subsection{User Study}
To assess the effectiveness of our method, we conducted a user study comparing our results with those of Animate Anyone~\cite{hu2024animateanyone} and UniAnimate~\cite{wang2024unianimate}, incorporating our environment in both baselines for fair comparison. Participants were asked to evaluate videos based on four metrics: 1) Character Consistency (CC), measuring the preservation of character identity and deformation; 2) Motion Fidelity (MF), assessing alignment of generated motion with the original; 3) Camera Movement Alignment (CMA), evaluating consistency with the original camera trajectory; and 4) Overall Harmony (OH), reflecting the overall visual appeal and naturalness. The study involved 120 participants who rated 64 videos randomly selected from 30 shots and 48 generated characters. Each participant reviewed 12 randomly chosen videos per method and compared them to the corresponding original shots. 

As shown in Table~\ref{tab:user_study}, our method consistently outperformed both baselines across all four metrics. In particular, the improvement in \textbf{Character Consistency} and \textbf{Motion Fidelity} highlights the benefit of structure-guided animation in preserving body shape and dynamics, even under complex poses or rapid movements. Our method also achieved significantly higher \textbf{Camera Movement Alignment} scores, reflecting the effectiveness of shape-aware camera optimization in producing accurate cinematic framing and avoiding motion drift. Furthermore, the elevated \textbf{Overall Harmony} scores demonstrate that our environment-aware generative refinement successfully reduces lighting and stylistic mismatches, yielding videos with greater aesthetic appeal and fewer perceptual artifacts.

\section{Conclusion}
\label{sec:conclusion}
In this paper, we introduce \textbf{DreamCinema}, a novel framework for simplifying the process and making film creation more accessible. Our key insight is to decompose film shots into four components (\ie, 3D character, driven motion, camera movement, and environment) and model them in 3D space, which naturally preserves 3D consistency and enables flexible subsequent applications. However, due to the arbitrary nature of these components, they are not always perfectly aligned, leading to issues during the reproduction. We further propose structure-guided character animation, shape-aware camera movement optimization, and environment-aware generative refinement, which allow us to recreate high-quality film shots with 3D consistency, high-fidelity motion, diverse camera movements, and overall harmony. Furthermore, we demonstrate that our framework provides flexibility in manipulating all elements, offering a new path for everyone to be their own filmmaker.
While our current system focuses on character-centric shots, it can be easily extended with more general pose and motion extraction methods to support a broader range of scenes. We look forward to integrating a more general camera pose estimation method and motion extraction approach suitable for a wider range of objects within our framework.


\section*{Acknowledgment}

This work was supported in part by the National Natural Science Foundation of China under Grant 62206147.

\ifCLASSOPTIONcaptionsoff
  \newpage
\fi

\bibliographystyle{IEEEtran}
\bibliography{bare_jrnl}

\end{document}